\documentclass[runningheads]{llncs}
\usepackage{graphicx}
\usepackage{amsmath,amssymb} 
\usepackage{color}
\usepackage[width=122mm,left=12mm,paperwidth=146mm,height=193mm,top=12mm,paperheight=217mm]{geometry}

\usepackage{caption} 
\usepackage{multirow}

\begin{document}
\pagestyle{headings}
\mainmatter

\title{Recurrent Neural Network for Learning Dense Depth and Ego-Motion from Video} 

\titlerunning{Reccurrent}

\authorrunning{authors running}

\author{Rui Wang, Jan-Michael Frahm, Stephen M. Pizer}

\institute{Computer Science, \\
          University of North Carolina at Chapel Hill\\
          \email{\{wrlife,jmf,smp\}@cs.unc.edu}
}

\maketitle

\begin{abstract}

Learning-based, single-view depth estimation often generalizes poorly to unseen datasets. While learning-based, two-frame depth estimation solves this problem to some extent by learning to match features across frames, it performs poorly at large depth where the uncertainty is high. There exists few learning-based, multi-view depth estimation methods. In this paper, we present a learning-based, multi-view dense depth map and ego-motion estimation method that uses Recurrent Neural Networks (RNN). Our model is designed for 3D reconstruction from video where the input frames are temporally correlated. It is generalizable to single- or two-view dense depth estimation. Compared to recent single- or two-view CNN-based depth estimation methods, our model leverages more views and achieves more accurate results, especially at large distances. Our method produces superior results to the state-of-the-art learning-based, single- or two-view depth estimation methods on both indoor and outdoor benchmark datasets. We also demonstrate that our method can even work on extremely difficult sequences, such as endoscopic video, where none of the assumptions (static scene, constant lighting, Lambertian reflection, etc.)  from traditional 3D reconstruction methods hold.

\end{abstract}

\section{Introduction}

The task of 3D reconstruction from monocular video is a longstanding task in computer vision. The state-of-the-art algorithm for dense monocular 3D reconstruction involves steps including Simultaneous Localization and Mapping (SLAM) or Structure from Motion (SfM) to get a semi-dense or sparse 3D reconstruction and camera pose estimates. Subsequently multi-view stereo (MVS) methods are used to get dense 3D reconstructions. Despite the progress in current visual SLAM \cite{LSDSLAM,ORBSLAM,DTAM}, SfM \cite{COLMAP,wu2013towards,agarwal2011building,frahm2010building} and MVS algorithms \cite{schonberger2016pixelwise,furukawa2010towards,furukawa2010accurate}, this reconstruction pipeline still has some inherent limitations; it can only work in static scenes with rigid objects; it requires a sufficient motion baseline for the cameras; and it assumes static lighting condition and Lambertian surface reflection. Our algorithm addresses all three limitations to enable highly flexible, video-based joint camera motion and dense geometry estimation. 



Recently, convolutional neural networks (CNN) \cite{liu2015deep,eigen2015predicting,garg2016unsupervised,zhou2017unsupervised,ummenhofer2017demon} have began to produce results of comparable quality to traditional geometric computer vision methods for depth estimation. However, most methods can take only a single frame or pair of frames as input, or report no benefit from additional frames. For example, Zhou \textit{et al.} \cite{zhou2017unsupervised} report that adding more frames for their technique does not improve the estimation accuracy, as their CNN can only capture the spatial relationships of the input. When their network receives stacked images as input, the temporal ordering is lost.  


\begin{figure}[h]
    \centering
    \includegraphics[width=12.5cm]{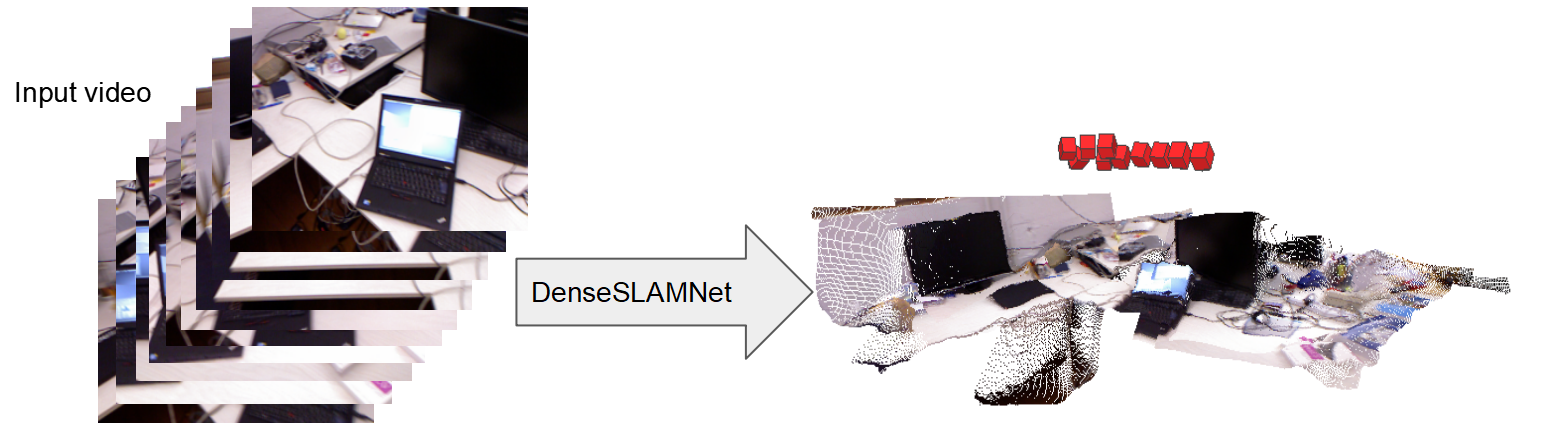}
    \caption{Our proposed DenseSLAMNet takes successive video frames as input and outputs a high quality depth map and camera pose for every input frame.}
\end{figure}

We have developed a recurrent neural network (RNN) for dense visual SLAM that simultaneously estimates the camera poses and dense depth maps from a video sequence taken by a monocular camera. In an RNN, the input to each layer includes information about the previous prediction, and thus explicitly takes temporal information into account. As far as we know, this is the first learning-based dense SLAM method that can estimate camera motion and dense depth maps in an unconstrained multi-view environment. We have improved upon existing deep single- and two-view stereo depth estimation methods by interleaving Long Short-Term Memory (LSTM) units with convolutional layers to effectively utilize multiple previous frames in each estimated depth maps.

In this paper, we present DenseSLAMNet, a network that can sequentially estimate depths and camera motion from monocular video. Our primary innovation is to incorporate LSTM units, commonly used in natural language processing, into a depth estimation network. These LSTM units allow the depth and camera motion estimation to become a multi-view process. We evaluate our network on several 3D benchmark datasets (SUN3D, RGBD-SLAM, NYUDepthV2, KITTI, and Make3D) and real patient endoscopic data. We analyze the effectiveness of our method on both deformable and rigid scenes. We summarize our contributions as follows:
\begin{itemize}
    \item We introduce a new RNN architecture for depth estimation from multiple views.
    \item We show that our multi-view depth estimation outperforms existing single-view methods.
    \item We demonstrate the successful application of our framework on endoscopic videos, a particularly challenging data modality for depth estimation.
\end{itemize}

\section{Related work}

SfM and SLAM are the two most prevalent frameworks for sparse 3D reconstruction of rigid geometry from images. SfM is typically used for offline 3D reconstruction from unordered image collections, while visual SLAM aims for a real-time solution using a single camera \cite{MONOSLAM,PTAM}. Sch\"{o}nberger and Frahm \cite{COLMAP} review the state-of-the-art in SfM and propose an improved incremental SfM method. More recent works on sparse SLAM systems include ORB-SLAM \cite{ORBSLAM} and DSO \cite{DSO}.  

While sparse methods use detected feature points for reconstruction, dense (or semi-dense) methods attempt to reconstruct all pixels from the 2D image. LSD-SLAM \cite{LSDSLAM} is a semi-dense SLAM method that operates directly on image intensities both for tracking and mapping. The DTAM framework \cite{DTAM} creates a dense 3D surface model through direct dense image registration and immediately uses it for camera tracking. Our DenseSLAMNet falls into this category of dense reconstruction methods. Multi-view stereo (MVS) \cite{schonberger2016pixelwise,furukawa2010towards,furukawa2010accurate} is another dense reconstruction method that generates dense depth maps using camera poses and raw image data. Frequently, MVS is used together with SfM. All above sparse and dense reconstruction methods require a static scene, constant illumination, and sufficient camera motion baseline for accurate reconstruction. In this paper, we present a method that can perform single- and multi-view dense 3D reconstruction for both static and deformable scenes with either constant or inconsistent light conditions.

Recently, researchers have started to apply CNNs to the 3D reconstruction problem. Eigen \textit{et al.} \cite{eigen2015predicting} and Liu \textit{et al.} \cite{liu2015deep} propose end-to-end networks, while other work has used CNNs for components of the pipeline, including correspondence matching \cite{yi2016lift,ilg2017flownet},
camera pose estimation \cite{kendall2015posenet}, and stereo \cite{luo2016efficient,kendall2017end}. Common output representations include depth maps, point clouds, and voxels. The advantage of these learning-based methods over the classical SfM-MVS pipeline is that we can leverage semantic supervision during the training process. This can lead to better reconstructions of texture-less or occluded surfaces and very thin structures, both of which are challenging for purely geometric techniques. 

A particular case of dense geometry estimation is monocular depth estimation. Monocular depth estimation has gained interest because regressing the depth representation is similar to the segmentation problem and thus the structure of CNNs can be easily adapted to the task of depth estimation \cite{FCN}. Eigen \textit{et al} \cite{eigen2015predicting} proposed an early multi-scale, end-to-end, per-pixel depth estimation framework. Laina \textit{et al}. \cite{laina2016deeper} extended Eigen’s work with a deeper residual network. More recently, incorporating elements of view synthesis \cite{zhou2016view} and Spatial Transform Networks \cite{jaderberg2015spatial}, Gordard \textit{et al}. \cite{godard2017unsupervised}, Garg \textit{et al}. \cite{garg2016unsupervised}, Zhou \textit{et al}. \cite{zhou2017unsupervised}, have trained end-to-end monocular depth estimation networks without ground-truth. This was done by transforming the depth estimation problem into an image reconstruction problem where the depth is the intermediate product that integrates into the image reconstruction loss. Despite the fact that these unsupervised depth estimation methods eliminate the complication of obtaining ground-truth depth, none outperform the traditional SfM or SLAM methods \cite{COLMAP,agarwal2011building,DTAM,LSDSLAM}.

Two-view or multi-view stereo methods have traditionally been the most common techniques for dense depth estimation. For the interested reader, Scharstein and Szeliski \cite{scharstein2002taxonomy} give a comprehensive review on two-view stereo methods. Newcombe \textit{et al.} \cite{DTAM} demonstrates that estimated depth accuracy becomes more precise as the number of views increases, even with small baseline motion. We leverage this result, explicitly learning correspondences between nearby frames which results in a similar multi-view benefit.  Recently, Ummenhofer \textit{et al}. \cite{ummenhofer2017demon} formulated two-view stereo as a learning problem. They showed that by explicitly incorporating dense correspondences estimated from optical flow into the two-view depth estimation, they can force the network to utilize stereo information on top of the single view priors. There is currently a very limited body of CNN based multi-view reconstruction methods. Choy \textit{et al}. \cite{choy20163d} use an RNN to reconstruction the object in the form of a 3D occupancy grid from multiple viewpoints. Rezende \textit{et al}. \cite{rezende2016unsupervised} introduced a family of generative models of 3D structures and recover these structures from 2D images via probabilistic inference. They learn the complex 3D to 2D projection through a generative model in an unsupervised way. However, these methods target single object reconstruction and fail for deformable objects.

\begin{figure}[h]
    \centering
    \includegraphics[width=11cm]{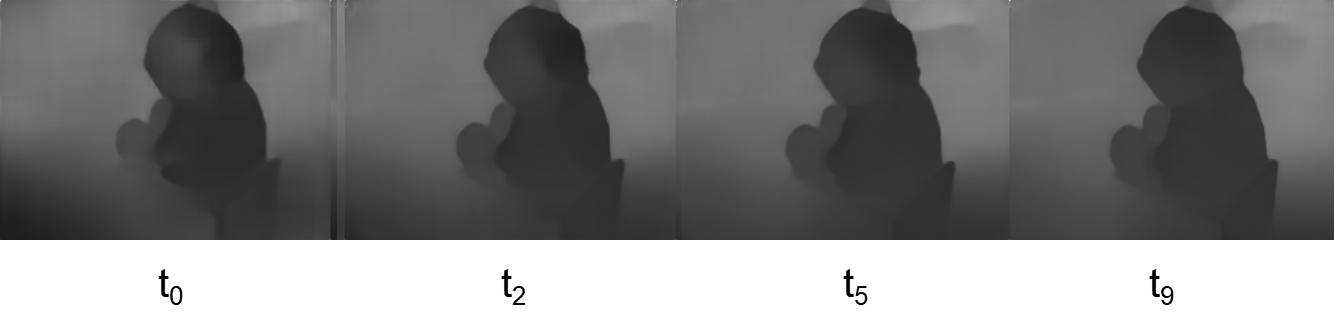}
    \caption{An example of the small-baseline refinement using LSTM layers.}
    \label{refine}
\end{figure}

Our approach is most closely related to dense visual SLAM in that camera motion and depth maps are estimated from multiple views in a sequential manner.  Tanteno \textit{et al}.\cite{tateno2017cnn} proposed CNN-SLAM, which predicts a depth map as an initial guess and subsequently refines it with a direct SLAM scheme relying on small-baseline stereo matching. Our DenseSLAMNet, as shown in Figure \ref{refine}, implicitly performs the small-baseline refinement via the information preserved across time-steps by the hidden layers of the LSTM. 

\section{Network architecture}

\begin{figure}[h]
    \centering
    \includegraphics[width=12cm]{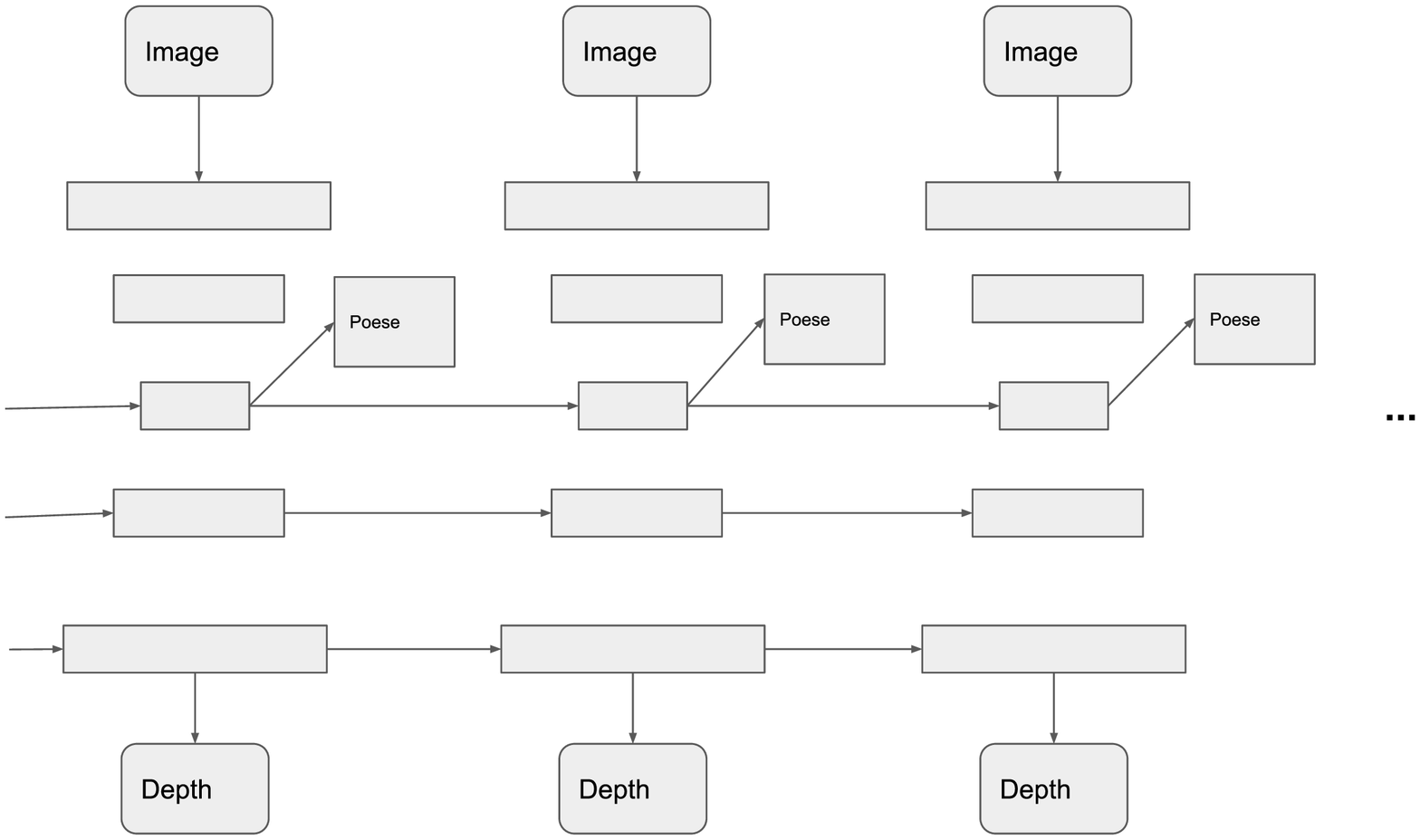}
    \caption{Overall network architecture of DenseSLAMNet.}
    \label{overall}
\end{figure}

Our DenseSLAMNet can simultaneously estimate dense depth maps and camera poses from a monocular video sequence under different scenarios (indoor, outdoor, endoscopy). We incorporate recurrent units into a CNN to leverage temporal information in our depth estimation, making it more accurate for continuous video sequences. However, unlike DeMoN \cite{ummenhofer2017demon} which is restricted to two-view input, our DenseSLAMNet takes a single frame at a time as input, but can operate over longer image sequences. It can also perform single-view depth estimation when required. Although our network incorporates temporal information through the recurrent units, it operates on each individual frame independently during training. This is contrary to Zhou \textit{et al}. \cite{zhou2017unsupervised}, Godard \textit{et al}. \cite{godard2017unsupervised}, and dense SLAM methods \cite{LSDSLAM,ORBSLAM,DTAM} which utilize relative geometry between frames. Therefore, our method is not restricted to static scenes or constant scene illumination. Figure \ref{endo} (a) shows an example of method estimating depth from endoscopic videos, where the scene frequently deforms and the light source moves with the camera, changing the scene illumination throughout the video sequence.

The overall architecture of our network is shown in Figure \ref{overall}. It takes a single RGB frame $I_t$ and the hidden states $h_{t-1}$ from the previous time step as input. The hidden states are transmitted internally through the LSTM units. The output of our network is the depth map $z_t$ and the camera pose $\{R_t, T_t\}$ of the current frame. Similar to single-view depth estimation networks, our DenseSLAMNet takes only a single frame at a time as input. Therefore, our network can perform both single-frame and multi-frame depth estimation. This makes our DenseSLAMNet more flexible than both CNN-based single-view depth estimation methods and visual SLAM methods.



\begin{figure}[h]
    \centering
    \includegraphics[width=12.5cm]{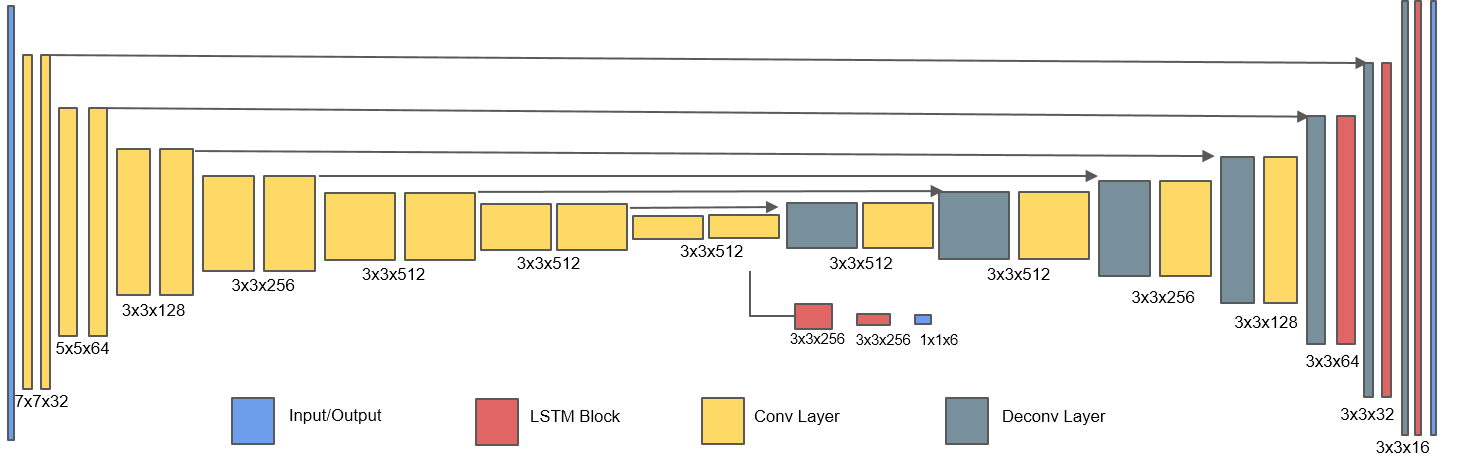}
    \caption{(Best viewed in color) Our network architecture at a single time step. We use the DispNet architecture. The width and height of each rectangular block indicates the size and the number of the feature map at that layer. Each increase and decrease of size represents a change factor of 2. The first convolutional layer has 32 feature maps. The kernel size for all convolution layers is 3, except for the first two convolution layers, which are 7 and 5, respectively.}
    \label{detail}
\end{figure}

Figure \ref{detail} shows our network at a single time step in more detail. Different colors encode the different units: yellow is a convolutional layer, red is an LSTM block, dark gray is a deconvolutional layer, and blue is and input/output layer. Our network uses a U-shape network architecture similar to  DispNet \cite{mayer2016large}. The height of each rectangle in Figure \ref{detail} represents the size of its feature maps, where each smaller feature map is half the size of the preceding feature map. The down-sampling from a previous layer to the next is done by a stride-2 convolution instead of max-pooling. The lines connecting corresponding layers in the encoder and decoder are skip-connections. 



We denote the size of our temporal window by $N$. In all experiments, we use $N=10$ as the length of our temporal sequence. Hence, the network in Figure \ref{detail} is replicated 10 times as shown in Figure \ref{overall}, with the temporal information being passed between the three LSTM blocks at each time-step.

\section{Training procedure}

\begin{figure}[h]
    \centering
    \includegraphics[width=12.5cm]{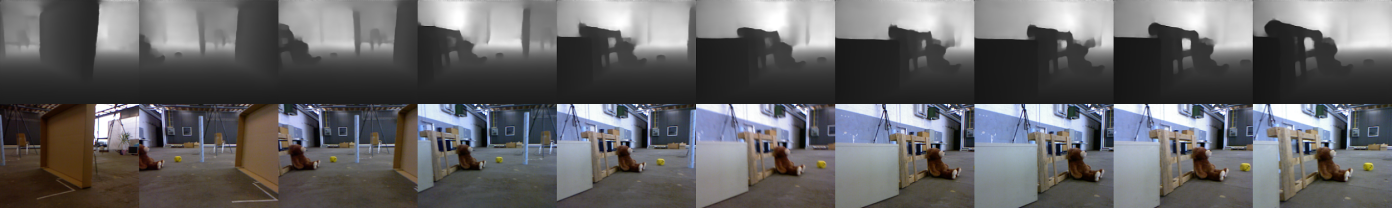}
    \caption{Example training data with a temporal window size of ten.}
    \label{training}
\end{figure}

For the ease of training and data preparation, we use a temporal window size of $N=10$, but ideally, similar to natural language processing, the network should take an arbitrary length sequence as input for training. During training, we feed frames to the network and compute losses from all frames in a temporal window. However, there is no input length constraint at test time. Even though the network can only store information from up to ten frames, longer sequences can still yield better results because the each prior frame has already been boosted by its previous ten frames. Figure \ref{training} shows a example of our training data.

\subsection{Loss function}

Our loss function is a composition of a point-wise depth loss, a camera pose loss, and a scale-invariant gradient loss. Similar to DeMoN \cite{ummenhofer2017demon}, we use disparity, the reciprocal of depth, $\xi=\frac{1}{z}$ as our direct estimation because it can represent points at infinity and account for the localization uncertainty of points at increasing distance. For camera pose, we use the Euler angle $R$ and the translation vector $T$. In total there are 6 parameters in the pose parameterization. 
Our point-wise depth loss is formulated as follows:
\begin{equation} 
L_{depth} = \sum_{t}^N\sum_{i,j}|\xi_t(i,j)-\hat{\xi_t}(i,j)|
\end{equation}
where $i,j$ is the pixel location in a depth map and $t$ represents the time-step. In this work, we use temporal window is $N=10$, so $t \in [0,9]$.
For depth we use an $L_1$ loss due to its robustness to noise. The overall depth loss integrates over all pixels as well as all frames in a temporal window.

To ensure the smoothness and sharpness of the estimated depth, we have adopted a loss on a scale-normalized gradient-like measurement, as introduced by Ummenhofer \textit{et al.} \cite{ummenhofer2017demon}. This loss is defined as
\begin{equation} 
L_{grad} = \sum_t\sum_{h\in\{1,2,4,8,16\}}\sum_{i,j}||g_{h,t}(i,j)-\hat{g}_{h,t}(i,j)||_2
\end{equation}
where $h$ is a spatial step size for computing $g_{h,t}$ at different scale. The vector $g_{h,t}$ is a scale-normalized, discretized measurement of the local changes of $\xi_t$. The measurement is defined as
\begin{equation} 
g_{h,t} = (\frac{\xi_t(i+h,j)-\xi_t(i,j)}{|\xi_t(i+h,j)|+|\xi_t(i,j)|}, \frac{\xi_t(i,j+h)-\xi_t(i,j)}{|\xi_t(i,j+h)|+|\xi_t(i,j)|})^T
\end{equation}

$L_{grad}$ in Eq. (2) emphasizes the depth discontinuities, such as occlusion boundaries and sharp edges, as well as the smoothness in homogeneous regions. This property encourages the estimated depth map to preserve more details and reduce noise. Therefore, we put highly weight on this component of the loss.

For the camera poses, the losses are weighted separately for rotation $R$ and translation $T$.
\begin{equation}
\begin{aligned}
L_{rot} &= \sum_t||r_t-\hat{r_t}||_2\\
L_{trans} &= \sum_t||t_t-\hat{t_t}||_2
\end{aligned}
\end{equation}

The overall loss is a weighted sum of $L_{depth}$, $L_{grad}$, $L_{rot}$, and $L_{trans}$, where the weights are chosen empirically. 

\textbf{Training details.} We set the weights for depth loss, scale-invariant gradient loss, camera rotation loss, and camera translation loss to 500, 1000, 500, and 100, respectively. We use the Adam \cite{kingma2014adam} optimizer with $\beta_1=0.9$, $\beta_2=0.999$. The initial learning rate is 0.0002 and decays exponentially every 10,000 steps by a factor of 0.9. For indoor scenes and endoscopic data, we resized the images to 192$\times$256. For outdoor scenes we resize the images to 128$\times$416. The image sizes are chosen for both computational efficiency and to be consistent with existing methods. We trained and evaluated our network on indoor and outdoor scenes separately. Different datasets have different camera intrinsic parameters, so we explicitly crop and resize images to ensure uniform intrinsic parameters. This step assures that the non-linear mapping between color and depth is consistent across all training datasets.

\section{Experiments}
We evaluate our method on multiple datasets and compare against the state-of-the-art for learning-based, depth estimation methods. 

\subsection{Training datasets}

\textbf{Indoor}. We use two publicly available datasets for indoor scenes. The first one is \textbf{SUN3D} \cite{xiao2013sun3d}, which is a large dataset with ground truth depth maps and camera poses. We selected 192 scenes from a total of 354 scenes as our training data. Then we randomly selected 30 scenes from the remaining 162 scenes as for validation and testing. The second dataset is \textbf{RGBD-SLAM} \cite{sturm12iros}, which is a smaller dataset but with higher camera pose accuracy. RGBD-SLAM provides a training and validation split of their dataset, so we directly use their split. In addition, we used the \textbf{NYUDV2} \cite{Silberman:ECCV12} dataset for generalization evaluation.

\textbf{Outdoor}. We use the KITTI dataset \cite{Geiger2013IJRR} for outdoor scenes. To perform a consistent comparison with existing methods, we used the \textbf{Eigen Split} \cite{eigen2015predicting} to train and evaluate our network. The \textbf{Make3D} \cite{saxena2009make3d} dataset is used to evaluate generalization.

\textbf{Endoscopy} (challenge dataset). We also explore the 3D reconstruction of humans' inner body surfaces from endoscopic videos. For qualitative evaluation and training, we generated an endoscopic dataset containing 65,235 frames of video from 16 patients. We generate depth maps and camera poses using the SFMS method \cite{wang2017improving}. We train our model on 14 patients and test on 2 patients.

\subsection{Evaluation metrics}

At test time, our DenseSLAMNet runs in real-time, at approximately 40 frames-per-second on a machine with a GeForce GTX1080 GPU. 

We evaluate DenseSLAMNet using five error metrics:

\begin{equation}
sc-inv(z,\Hat{z})  = \sqrt{\frac{1}{n}\sum_id^2_i-\frac{1}{n^2}(\sum_id_i)^2}
\end{equation}
where $d_i=log_{10}(z_i)-log_{10}(\Hat{z_i})$. $Sc-inv$ is a scale invariant error \cite{eigen2015predicting} that can evaluate depth regardless of scale.

\begin{equation}
Abs-rel(z,\Hat{z}) = \frac{1}{n}\sum_i\frac{|z_i-\Hat{z_i}|}{\hat{z_i}},
Abs-inv(z,\Hat{z}) = \frac{1}{n}\sum_i|\frac{1}{z_i}-\frac{1}{\Hat{z_i}}|
\end{equation}
$Abs-rel$ measures the relative difference of output predictions and the ground truth depth. It emphasizes close objects in the ground truth.
$Abs-inv$ also measures this relative difference, but even further emphasizes closer objects.
\begin{equation}
RMSE(z,\Hat{z})  = \sqrt{\frac{1}{n}\sum_i(z_i-\Hat{z_i})^2},
RMSE-log(z,\Hat{z})  = \sqrt{\frac{1}{n}\sum_i(d_i)^2}
\end{equation}
$RMSE$ and $RMSE-log$ are two of the most commonly used error measurements \cite{eigen2015predicting,zhou2017unsupervised,godard2017unsupervised,garg2016unsupervised,kuznietsov2017semi}, the first measuring absolute depth error and the second measuring absolute log-depth error. 

\subsection{Comparison with Existing Methods}\label{sec:compare}

\begin{figure}[h]
    \centering
    \includegraphics[width=12.5cm]{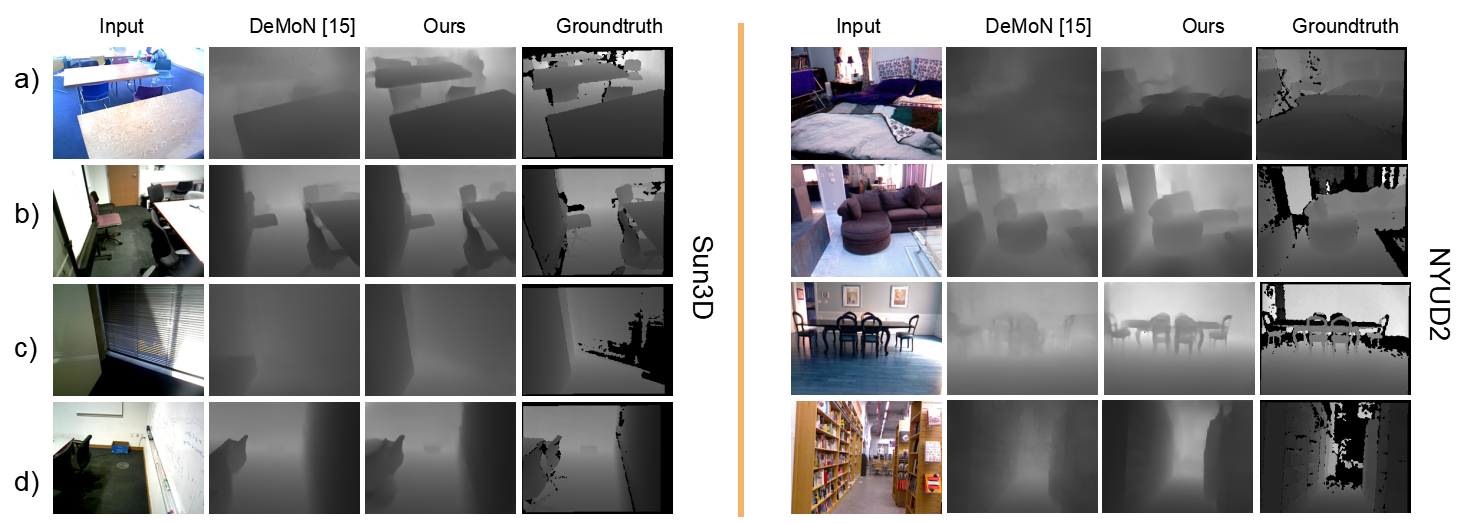}
    \caption{Visual comparison of ours vs. DeMoN's results \cite{ummenhofer2017demon} on SUN3D \cite{xiao2013sun3d} dataset. As can be seen in row (a) and (d), our DenseSLAMNet performs better at large distance.}
    \label{compare}
\end{figure}

We compared our DenseSLAMNet to state-of-the-art CNN-based, single- and two-view depth estimation methods. 

We compared to Eigen \textit{et al.} \cite{eigen2015predicting}, Liu \textit{et al.} \cite{liu2015deep}, and DeMoN \cite{ummenhofer2017demon}. Eigen \textit{et al.} and Liu \textit{et al.} are single-frame depth estimation methods, and DeMoN is a two-view depth estimation method. We take their publicly available pre-trained models and test on our prepared testing data. Eigen \textit{et al.} and Liu \textit{et al.} methods are trained on NYUDV2 dataset. DeMoN is trained on several indoor, outdoor, and synthetic datasets, including SUN3D and RGBD-SLAM. In order to evaluate the full capability of our network on single-frame depth estimation, we feed DenseSLAMNet video sequences of size $N=10$ during testing and report results on the last frame. When evaluating against DeMoN, we report their result as the pair of frames within each temporal window that gives best score for their method. Table \ref{table:1} shows that our method outperforms the existing state-of-the-art across every quantitative metric for indoor scenes. 

\begin{figure}[h]
    \centering
    \includegraphics[width=10cm]{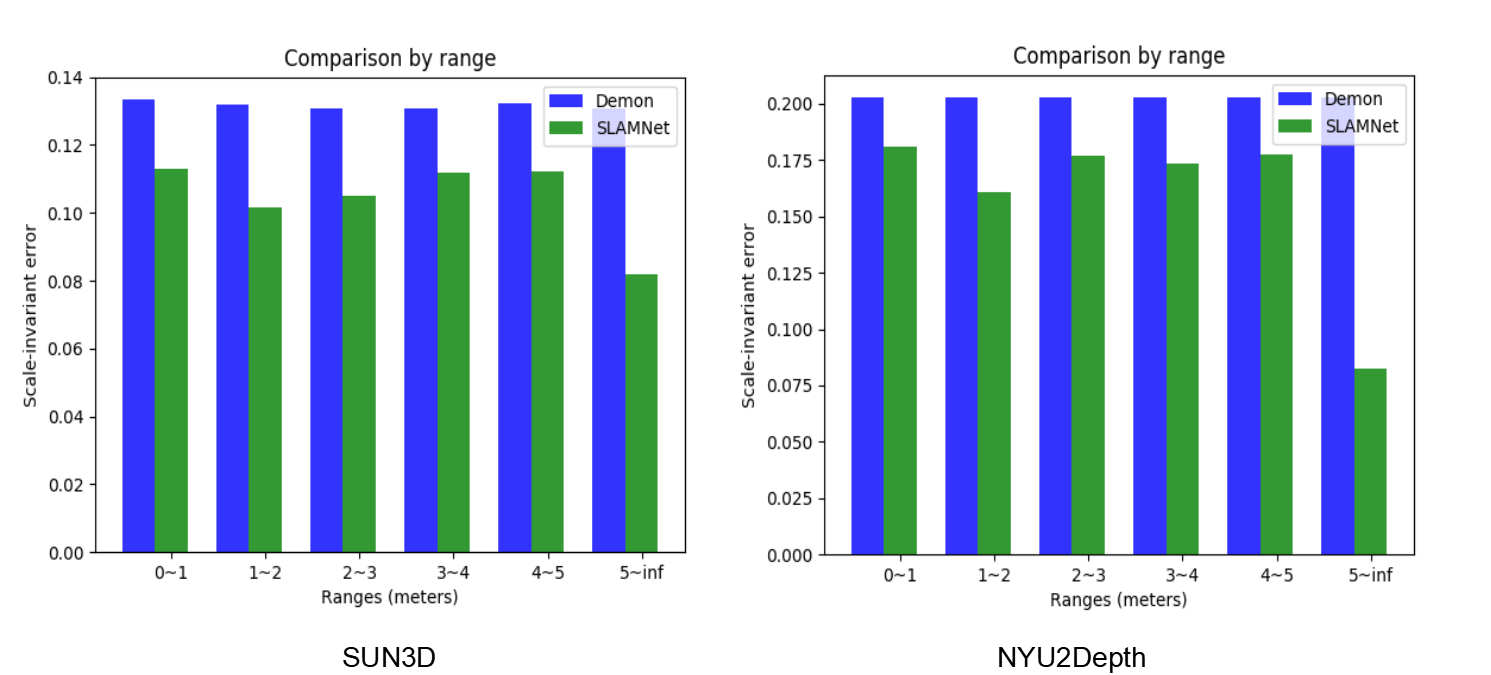}
    \caption{Comparison with DeMoN \cite{ummenhofer2017demon} at different ranges. We outperform DeMoN in all ranges, especially at large distance.}
    \label{range}
\end{figure}

Figure \ref{compare} shows a visual comparison of our DenseSLAMNet with other methods. It can be seen that our DenseSLAMNet produces sharper results than DeMoN, the second best performing method. Rows (a) and (d) in Figure \ref{compare} also demonstrate that we perform significantly better for larger distances. Figure \ref{range} shows a detailed comparison between DeMoN and our method at different depth ranges. We measure the average $sc-inv$ error at different depth ranges, eg. 0 to 1 meters, 1 to 2 meters and so on, across all testing images. As can be seen, our method consistently performs better at all ranges and especially at large distances. 

\begin{table}[h!]
\centering
\begin{tabular}{ |p{3cm}||p{2cm}|p{2cm}|p{2cm}| }
 \hline
 Methods & Sc-inv & Abs-inv & Abs-rel \\
 \hline
 Eigen \textit{et al.} \cite{eigen2015predicting} &0.190 & 0.068&  0.177\\
 Liu \textit{et al.} \cite{liu2015deep} &0.215 & 0.070&  0.212\\
 DeMoN \cite{ummenhofer2017demon} &0.128 & 0.047&  0.109\\
 \textbf{DenseSLAMNet} & \textbf{0.112} & \textbf{0.039}&  \textbf{0.091}\\ 
 \hline
\end{tabular}
\caption{Quantitative comparison of our DenseSLAMNet with the state-of-the-art CNN-based methods on SUN3D \cite{xiao2013sun3d} dataset. Lower numbers are better.}
\label{table:1}
\end{table}

\begin{table}[h!]
\centering
\begin{tabular}{ |p{3cm}||p{1.2cm}||p{1.5cm}|p{1.5cm}|p{1.5cm}|p{1.5cm}|  }
 \hline
 Methods & Dataset & Abs-rel & Sq-rel & RMSE& RMSE-log\\
 \hline
 Eigen \textit{et al}. \cite{eigen2015predicting} & K  & 0.203  & 1.548 & 6.307 & 0.282 \\
 Liu \textit{et al}. \cite{liu2015deep} & K &   0.202  &  1.614  & 6.523 & 0.275 \\
 Kuznietsov \textit{et al.} \cite{kuznietsov2017semi} & K & 0.113 & 0.741 & \textbf{4.621} & 0.189\\
 Zhou \textit{et al.} \cite{zhou2017unsupervised} & CS+K & 0.198 & 1.836 & 6.565 & 0.275 \\
 Godard \textit{et al.} \cite{godard2017unsupervised} & CS+K & \textbf{0.097} & 0.896 &  5.093 & \textbf{0.176} \\
 \textbf{DenseSLAMNet} & K & 0.129 & \textbf{0.704}&  4.743 & 0.199 \\ 
 \hline
 \hline
 \textbf{DenseSLAMNet} & K & \textbf{0.058} & \textbf{0.205}&  \textbf{2.538}& \textbf{0.087} \\
 \hline
\end{tabular}
\caption{Quantitative comparison of our DenseSLAMNet with other state-of-the-art CNN-based methods on KITTI \cite{Geiger2013IJRR} dataset using the Eigen Split \cite{eigen2015predicting}. The last row is our performance on continuous sequences. Lower numbers are better. K and CS stand for KITTI and Cityscapes, \cite{Cordts2016Cityscapes} respectively. All results are capped at 80m depth.}
\label{table:2}
\end{table}


\begin{figure}[h]
    \centering
    \includegraphics[width=12.5cm]{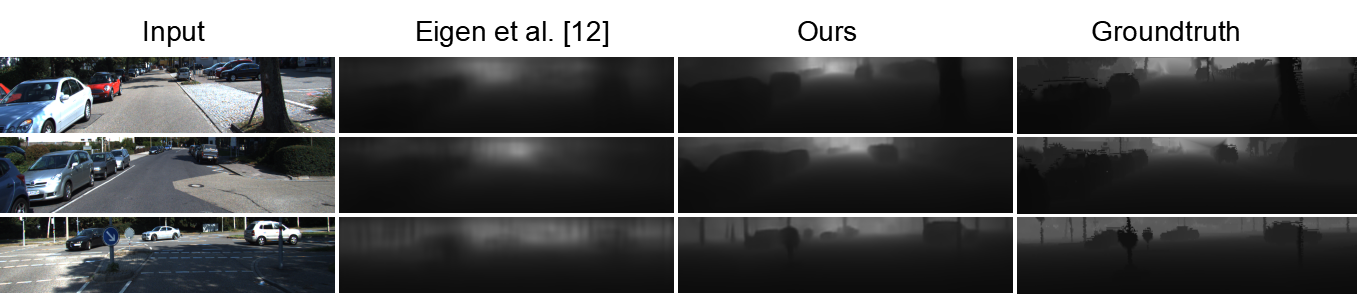}
    \caption{Visual comparison between the results of Eigen \textit{et al}.\cite{eigen2015predicting} and ours on KITTI dataset \cite{Geiger2013IJRR}. Groundtruth depth is interpolated for visualization purpose.}
    \label{KITTI}
\end{figure}

Table \ref{table:2} shows a quantitative comparison on outdoor scenes. We compare to Eigen \textit{et al}. \cite{eigen2015predicting}, Garg \textit{et al}. \cite{garg2016unsupervised}, Godard \textit{et al}. \cite{godard2017unsupervised}, Liu \textit{et al}. \cite{liu2015deep}, Zhou \textit{et al.} \cite{zhou2017unsupervised}, and Kuznietsov \textit{et al.} \cite{kuznietsov2017semi}. To perform a consistent comparison to state-of-the-art methods, we use the 697 test images from the Eigen Split \cite{eigen2015predicting} for evaluation. However, these 697 images are randomly selected from 28 scenes and do not form a continuous sequence. Therefore, they do not fully demonstrate the capability of our network. Despite this fact, Table \ref{table:2} shows that our method performs similarly to the state-of-the-art on this test set. In the last row of the table, we show an evaluation result of our method on continuous sequences that we randomly selected from the KITTI test dataset. It can be seen that our network gets a significant performance boost when dealing with continuous sequences, which we explore in more depth in Section \ref{sec:ablation}.

\begin{figure}[h]
    \centering
    \includegraphics[width=12.5cm]{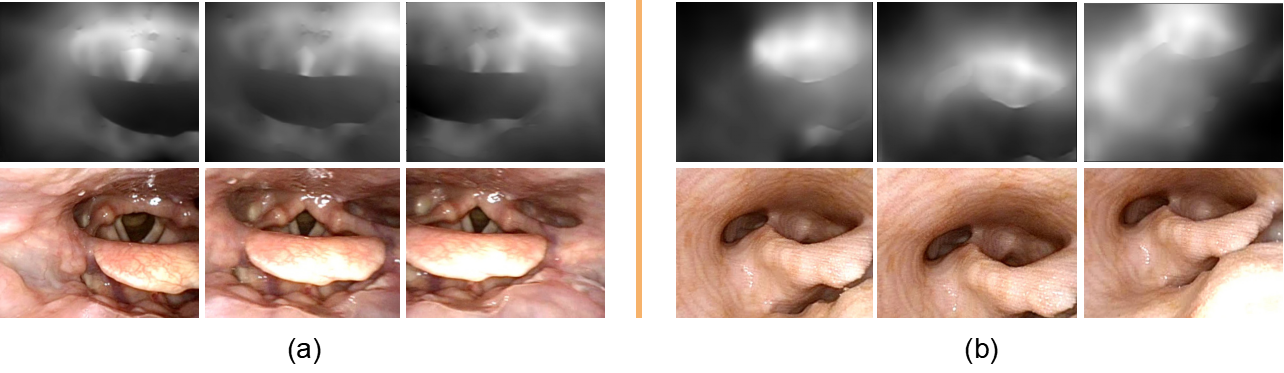}
    \caption{Visual results on real patient and phantom endoscopic data. The vocal cord is visibily closing in the image sequence in (a). In both datasets, one can see the illumination changes due to the moving light source.}
    \label{endo}
\end{figure}

We demonstrate DenseSLAMNet's ability to handle non-static scenes, moving light sources, and non-Lambertian surface reflections by training and testing it on an endoscopic dataset. Figure \ref{endo} shows a visual result of our DenseSLAMNet on this data modality. In order to perform a quantitative evaluation, we 3D printed a textured phantom throat model using geometry extracted from CT scans and performed an endoscopy process on it to capture video. DenseSLAMNet obtains 0.271 in $sc-inv$, 0.216 in $abs-rel$, and 0.010 in $abs-inv$. Figure \ref{endo} (b) shows a visual result on phantom dataset.


\subsection{Pose Estimation}

\begin{figure}[h]
    \centering
    \includegraphics[width=12.5cm]{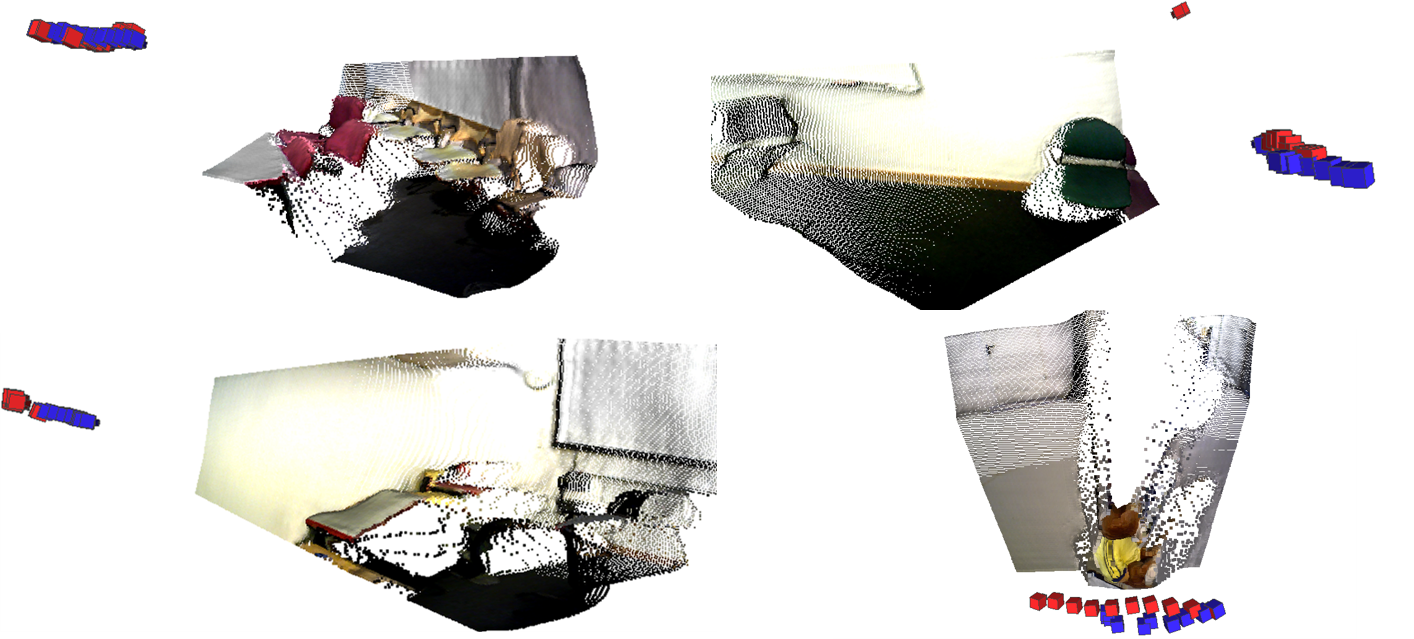}
    \caption{Camera pose estimation evaluation. Red cameras represent the ground truth camera positions and blue cameras represent our estimated camera positions. Here we only plot the point cloud of the last camera.}
    \label{cam}
\end{figure}

We provide a qualitative evaluation of the pose estimation task by plotting the predicted poses together with the ground truth camera poses. As can be seen from Figure \ref{cam}, our DenseSLAMNet can handle smooth and small camera motions very well, but fails on large sudden jumps and random camera motions. This is expected because, in the training data, the camera motions are small and smooth causing the network adapts to this specific type of camera motion. To handle random camera motion and large magnitude motion between frames, we suspect we might apply weights to different types of camera motion, or explicitly generate a balanced set of different types of camera motion sequences for training. We see this as an opportunity for future work.

\subsection{Generalization to new data}

\begin{table}[h!]
\centering
\begin{tabular}{ |p{3cm}||p{2cm}|p{2cm}|p{2cm}|  }
 \hline
 Methods & Sc-inv & Abs-inv & Abs-rel\\
 \hline
 DeMoN \cite{ummenhofer2017demon} &0.203 & 0.079&  0.201\\
 \textbf{DenseSLAMNet} & \textbf{0.181} & \textbf{0.071}&  \textbf{0.171}\\ 
 \hline
\end{tabular}
\caption{Generalization comparison between the depth estimation of DeMoN and our DenseSLAMNet on NYUDV2 dataset \cite{Silberman:ECCV12}.}
\label{table:3}
\end{table}

We evaluate the generalization ability of our DenseSLAMNet on both indoor and outdoor scenes. For indoor scenes, we use the NYUDV2 \cite{Silberman:ECCV12} dataset. The NYUDV2 does not provide the ground truth camera poses, so we cannot use it for training. Table \ref{table:3} shows the quantitative comparison results, and Figure \ref{compare} shows the visual results. Again our method outperforms DeMoN across every quantitative metric.

\begin{table}[h!]
\centering
\begin{tabular}{ |p{3.5cm}||p{2cm}|p{2cm}|p{2cm}|p{2cm}|}
 \hline
 Methods & Sq-rel & Abs-rel & RMSE& $log_{10}$ \\
 \hline
 Godard \textit{et al}. \cite{godard2017unsupervised} &11.990 & 0.535&  11.513& 0.156\\
 zhou \textit{et al}. \cite{zhou2017unsupervised} &5.321 & 0.383&  10.47& 0.478 \\
 Kuzenietsov \textit{et al}. \cite{kuznietsov2017semi} & - & 0.421&  8.237& 0.190\\
 \textbf{DenseSLAMNet} & \textbf{2.404} & \textbf{0.275}&  \textbf{6.476}& \textbf{0.102}\\ 
 \hline
\end{tabular}
\caption{Generalization comparison on Make3D dataset \cite{saxena2009make3d}. All results are capped at 70 meters depth.}
\label{table:4}
\end{table}

\begin{figure}[h]
    \centering
    \includegraphics[width=12.5cm]{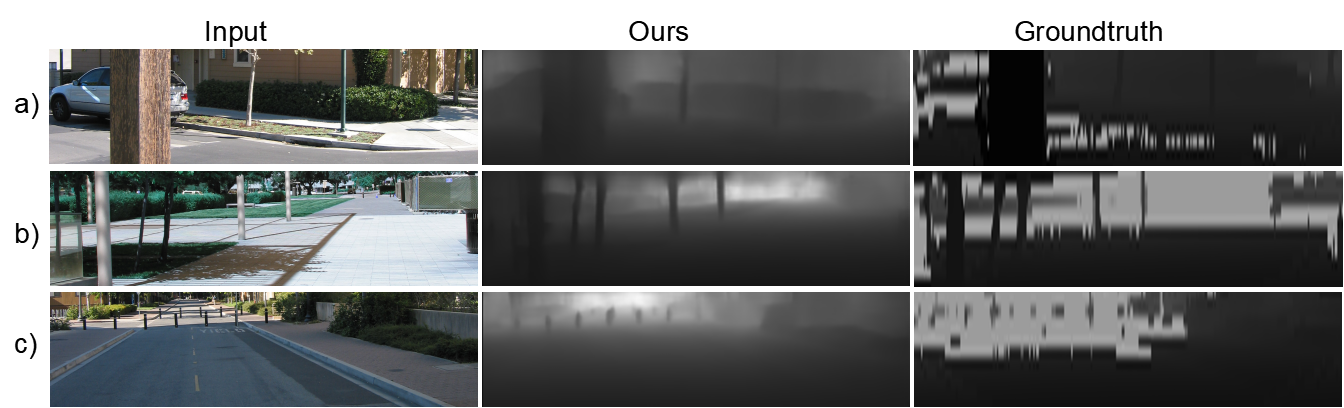}
    \caption{Our prediction on unseen outdoor dataset (Make3D).}
    \label{make3d}
\end{figure}

For outdoor scenes, we evaluate our DenseSLAMNet on the unseen Make3D dataset \cite{saxena2009make3d}. The Make3D dataset is very different from KITTI (used for training) in that its image resolution is 2272$\times$1702, whereas KITTI's is 375$\times$1280. Table \ref{table:4} shows the quantitative comparison results. Our method generalizes best among the state-of-the-art methods that are not trained on the Make3D dataset. From Figure \ref{make3d} we can see that our predicted depth map preserves fine details like trees, cars, and pillars.

\subsection{Ablation studies}\label{sec:ablation}
In order to justify the effectiveness of the different components of our network architecture, we did a series of ablation studies using the test dataset of SUN3D. 

\begin{table}[h!]
\centering
\begin{tabular}{ |p{3cm}||p{2cm}|p{2cm}|p{2cm} | }
 \hline
 Methods & Sc-inv & Abs-inv & Abs-rel \\
 \hline
CNN-SINGLE  & 0.131    & 0.049 & 0.103  \\
 CNN-STACK &   0.144  & 0.060   & 0.124 \\
 \textbf{DenseSLAMNet} & \textbf{0.112} & \textbf{0.039}&  \textbf{0.091}\\ 
 \hline
\end{tabular}
\caption{The use of LSTM in DenseSLAMNet gives the best depth estimation accuracy. Simply stacking up frames for training actually leads to worse performance than using just a single frame.}
\label{table:5}
\end{table}

We compared 3 types of networks. We trained a CNN-SINGLE network that uses the network architecture in Figure \ref{detail} but without LSTM (RNN) units. Then we trained another network, CNN-STACK, that uses the same network architecture as CNN-SINGLE. Instead of taking a single image as input, CNN-STACK takes a stack of ten images as input. Table \ref{table:5} shows the quantitative results of our analysis. These results demonstrate that the LSTM units make an important contribution to preserving temporal information across a video sequence, which leads to better depth maps.

\section{Conclusions}

In this paper, we presented a real-time, RNN-based, multi-view dense SLAM method for depth and camera pose estimation from single or multiple frames. Our method effectively utilizes the temporal relationships between neighboring frames through LSTM units, which we show is more effective than simply stacking multiple frames together as input. Our DenseSLAMNet outperformed nearly all of the state-of-the-art CNN-based, single-frame depth estimation methods on both indoor and outdoor scenes and showed better generalization ability. It also predicted more accurate depth at large distance compared to the existing state-of-the-art. In addition, we demonstrated its capability to estimate depth from especially difficult data: endoscopic videos with dynamic scene geometry and illumination. In the future, we would like to further investigate the camera pose estimation component to make our network robust to highly varied camera motion, as well as explore the possibility of training on variable-length temporal sequences.


\clearpage

\bibliographystyle{splncs}
\bibliography{egbib}

\end{document}